\documentclass[sigconf]{acmart}

\AtBeginDocument{%
  }

\setcopyright{acmcopyright}
\copyrightyear{2022}
\acmYear{2022}
\acmDOI{XXXXXXX.XXXXXXX}

\acmConference[Conference acronym 'XX]{Make sure to enter the correct
  conference title from your rights confirmation emai}{2022}{}
\acmPrice{15.00}
\acmISBN{978-1-4503-XXXX-X/18/06}

\usepackage{xspace}

\begin{document}


\def\name{{CBO}\xspace}
\title{Automated Compliance Blueprint Optimization with Artificial Intelligence}

\author{Abdulhamid Adebayo}
\email{hamid.adebayo@ibm.com}
\affiliation{%
  \institution{IBM T.J. Watson Research Center}
  \city{Yorktown Heights}
  \state{New York}
  \country{USA}
}

\author{Daby Sow}
\email{sowdaby@us.ibm.com}
\affiliation{%
  \institution{IBM T.J. Watson Research Center}
  \city{Yorktown Heights}
  \state{New York}
  \country{USA}
}

\author{Muhammed Fatih Bulut}
\email{mfbulut@us.ibm.com}
\affiliation{%
  \institution{IBM T.J. Watson Research Center}
  \city{Yorktown Heights}
  \state{New York}
  \country{USA}
}

\renewcommand{\shortauthors}{Adebayo et al.}

\begin{abstract}

For highly regulated industries such as banking and healthcare, one of the major hindrances to the adoption of cloud computing is compliance with regulatory standards. This is a complex problem due to many regulatory and technical specification (techspec) documents that the companies need to comply with.
The critical problem is to establish the mapping between techspecs and regulation controls so that from day one, companies can comply with regulations with minimal effort. We demonstrate the practicality of an approach to automatically analyze regulatory standards using Artificial Intelligence (AI) techniques. We present early results to identify the mapping between techspecs and regulation controls, and discuss challenges that must be overcome for this solution to be fully practical.

\end{abstract}



\maketitle

\section{Introduction}
    

For years now, cloud computing has been proven to be the go-to-market strategy for enterprises. The staggering growth of the cloud market is expected to reach $\$832.1$ billion by 2025 \cite{newswire}. However, only $20\%$ of the mission-critical enterprise workloads and sensitive data have been deployed so far in cloud, and the rest are still running on-premises \cite{forbes}. One of the significant challenges to adopting cloud for this remaining $80\%$ of the applications is the need to constantly comply with changing regulations.  Particularly, businesses that operate in highly regulated industries such as finance, healthcare and defense must meet stringent compliance requirements.

For example, financial institutions need to be compliant with the Payment Card Industry Data Security Standard (PCI/DSS) \cite{pcidsswebsite}. In healthcare, companies need to be compliant with Health Insurance Portability and Accountability Act (HIPAA) \cite{hipaawebsite}. Since healthcare companies usually need to process payments, they also need to comply with PCI/DSS. On the other hand, regulations can be location-specific, for example, organizations handling personally identifiable information of European citizens must comply with General Data Protection Regulation (GDPR) \cite{gdprwebsite}, and the equivalence of that for California residents is California Consumer Privacy Act (CCPA) \cite{ccpawebsite}. It is not uncommon for enterprises to comply with several regulations at once to conduct business for a given country. 

In response to the increase in compliance requirements, several cloud providers have setup specific clouds to serve their highly regulated industry customers. For example, Microsoft's Trusted Cloud supports financial, healthcare and government customers. Likewise, AWS's GovCloud addresses stringent compliance requirements of United States Government customers, offering protective measures to sensitive data, ranging from Personally Identifiable Information, financial data, to patient medical records. With the hybrid cloud market set to reach 145 billion U.S. dollars in 2026, the burden is on an enterprise to have an understanding of their compliance posture across platforms and services. This visualization start with the mapping of techspecs to regulation controls.

Regulation controls can be implemented and governed by technical specifications (techspecs). Center for Information Security (CIS) benchmarks \cite{cis} and Security Technical Implementation Guides (STIGs) from US Department of Defense \cite{stigs} are major techspecs that enterprises follow to enforce regulation controls. Usually, mappings between regulation \emph{controls} and techspec \emph{checks} are manually constructed by Subject Matter Experts (SME) to prove compliance with regulations. This exercise requires familiarity with the regulation landscape along with the technical knowledge of security implementations. This rare combination of skills make compliance process a time-consuming and expensive task. With the constantly evolving regulation and techspecs landscape, a solution is needed to help SMEs perform mapping tasks more efficiently. 

\begin{figure}[h]
  \centering
  \includegraphics[width=0.9\linewidth]{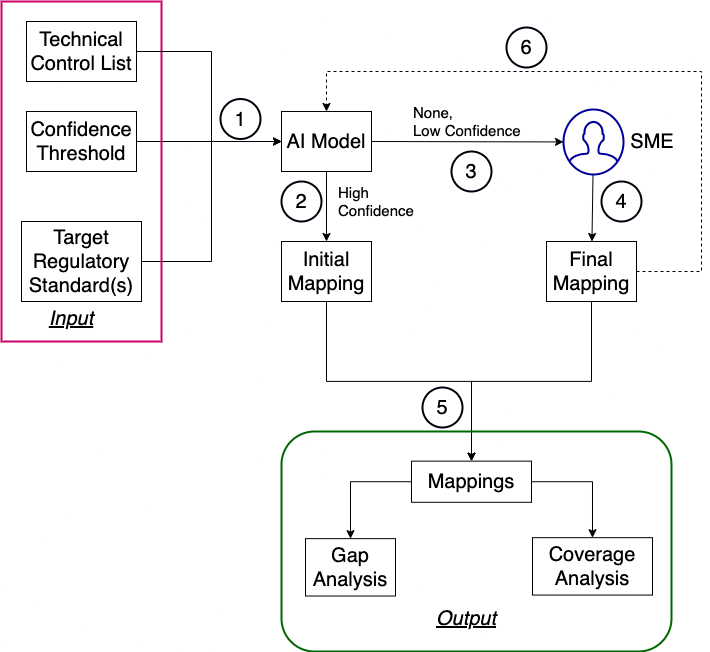}
  \caption{Overview of how our proposed approach improves enterprise compliance process.} 
  \label{fig:overview}
\end{figure}

In this paper, we outline a vision for an AI-assisted compliance blueprint optimizer (\name) -- see Figure \ref{fig:overview}, that will ease the compliance process for cloud-based systems by automatically:
\begin{itemize}
    \item mapping any given techspec text to a set of related regulation controls,
    \item incorporating SME feedback into the process to improve and provide more accurate mapping over time,
    \item providing an analysis of how what regulation controls have been covered based on the mappings (coverage analysis), and what outstanding controls need to be implemented with the techspecs (gap analysis).
\end{itemize}

\section{Motivation and Related Work}

It is pretty common for organizations to adopt multiple compliance frameworks (i.e., satisfy multiple regulatory standards) due to business requirements, geographical locations, or the domain of business functions. Security professionals then decode the controls specified in these regulations and implement systems, policies, and procedures to enforce security. However, there is often an overlap between the techspec checks and regulation controls. Similar relationships can also be found between controls of different regulations.

Regulatory documents and techspecs are often text-based. We consider mapping techspec checks to regulation controls as a supervised multilabel machine learning classification problem, where a single techspec check can be mapped to multiple regulation controls. In general, the multilabel problem has been addressed in the literature with many different approaches. [13] presented a simple neural network for multilabel text classification with the cross entropy error function. Long Short Term Memory (LSTM) \cite{zhou2015c, zhou2016text} and Convolutional Neural Networks (CNN) models \cite{hughes2017medical,chen2015convolutional,zhang2015character} have also been used for sentence and character level multilabel text classification. 


In \cite{chalkidis2019large}, the authors applied text classification approaches to EU legislation documents. Likewise, deep learning models \cite{liu2017deep, jabreel2019deep, you2018attentionxml} have also been applied to the problem of text classification across multiple domains. Specifically, pre-trained language models are an important part of AI solutions today, with Transformer models like BERT \cite{devlin2018bert}, RoBERTa \cite{liu2019roberta}, and ALBERT \cite{lan2019albert}. Pre-trained CNN-based models have also been found to be competitive and sometime outperform the Transformer models with convolutions faster and scale better to long sequences \cite{tay2021pre}. 

In \cite{adam2019cognitive}, the authors set out to match regulatory requirements with actual executable code that enforce the requirements. 
In \cite{agarwal2021ai}, authors presented an hierarchical classification approach for mapping customer's security controls to cloud provider's control set. Given a security control, \cite{agarwal2021ai} presented an hierarchical classification approach to mapping customer's security controls to cloud provider's control set. This approach  maps the security control to the family within the NIST framework, then attempts to map to the controls within the identified family. The mapping algorithm used is also optimized for high level customer controls rather than technical specifications, and will miss out on mappings to the leaf node in the hierarchy whose top-level family does not show enough correlation. 

Different from previous work, {\name} attempts to map a given description of a technical specification checks to a given regulation controls with active learning in place to continuously learn from mistakes and improve the accuracy.


\section{Mapping Problem}

Today, as state-of-the-art, techspec check to regulation controls mapping problem is treated as a text search problem, where a given text is compared against target regulation controls. 
For example, a user with a techspec check: \textit{Rule -  "Password expiration is set to 90 Days for existing passwords"} intends to map to \emph{NIST 800-53} controls. The user identifies the keywords as \emph{password}, \emph{expiration}, and \emph{days}. The \emph{NIST 800-53} publication is opened on a preferred file reader, and a manual search of the keywords is performed with the hope of finding the controls relevant to the techspec check. Our goal with {\name} is to reduce the latency of this workflow.
\begin{table*}[t]
\centering
    \begin{tabular}{|p{0.3\linewidth} | p{0.05\linewidth} | p{0.6\linewidth}|}
    \hline
        \textbf{Techspec checks} & \textbf{Target Regulation} & \textbf{Control Text}\\
    \hline\hline
    Check whether password policy requires at least one uppercase letter
 & NIST &   IA-5 (1)(b) Enforces at least the following number of changed characters when new passwords are created: [Assignment: organization-defined number] \newline IA-5(1) - AUTHENTICATOR MANAGEMENT | PASSWORD-BASED AUTHENTICATION\\
    \hline
    Ensure no more than 3 user administrators are defined for Kubernetes containers
 & NIST & AC-6 – LEAST PRIVILEGE | The organization employs the principle of least privilege, allowing only authorized accesses for users (or processes acting on behalf of users) which are necessary to accomplish assigned tasks in accordance with organizational missions and business functions.
\\

    \hline
     Check whether data disks are encrypted
 & HIPPA & 164.312(a)(2)(iv) Encryption and decryption - Implement a mechanism to encrypt and decrypt electronic protected health information.\\
    \hline
     Check whether data disks are encrypted
 & NIST & SC-28 - PROTECTION OF INFORMATION AT REST | The information system protects the [Selection (one or more): confidentiality; integrity] of [Assignment: organization-defined information at rest]. \newline SC-13 - CRYPTOGRAPHIC PROTECTION | The information system implements [Assignment: organization-defined cryptographic uses and type of cryptography required for each use] in accordance with applicable federal laws, Executive Orders, directives, policies, regulations, and standards.\\
 \hline
    \end{tabular}
\caption{Mapping Examples of Techspec checks to Regulation Controls.}
\label{tab:mapexamples}
\end{table*}


{\name} takes three inputs: 1) techspec text as a query, 2) the target regulatory standard (whose control set has been previously ingested to {\name}), and 3) threshold for the minimum similarity percentage between the query text and the predicted controls. The threshold allow the searcher to smoothly tradeoff prediction accuracy for granularity.
Next, we present details of the processes for preparing the training data for {\name}, the mapping algorithms used, how the result of the mapping process is presented, and how the SME feedback is continuously captured.

\subsection{Preprocessing} \label{sec:textPreprocess}
Regulatory standards are often expressed as text.
These documents are often available as spreadsheets, making them relatively easy to parse and process. First, the metadata relevant to the description of each control specification is identified. These metadata are title, rationale and remediate (fix).
Stop words and punctuation are removed, resulting in reduced data space and cleaner content. Additionally, the text is tokenized, normalized (all text converted to lower case), and de-noised (e.g., removing extra white spaces and unidentified characters). Each control specification is tagged with the associated regulation control identifier as the mapping.

\subsection{Text Search with Elasticsearch} \label{sec:search}
Elasticsearch (ES) provides a Lucene based search engine, exposed via APIs \cite{elastic}. The pre-processed text is batch-loaded into ES with each control specification and its associated meta-data represented as a document. 
Text is sent via the ES APIs to query a collection, and the response is a set of possible regulation controls relevant to the search query. Each result label has a corresponding relevance and confidence score. The relevance score is an unbounded measure of the relevance of a particular result, dependent on the query and matching document. A higher relevance score is indicative of a better match to the query parameters. The confidence score estimates how relevant a result is, with a value ranging from $0.0$ to $1.0$.

\subsection{Text Classification with CNN} \label{sec:classify}
Orthogonal to text search with ES, {\name} uses a second, machine learning based approach for mapping. Text classification uses a classifier to label unknown text given a pre-trained model. Text representation is an essential step in the training model development. The representation of the training data is vectorized before training the CNN model. Our implementation of the CNN algorithm is an adaptation of \cite{Kim14f, magpie, Berger2015LargeSM}. In a nutshell, the CNN model is a 3-layer Neural Network: the first layer is a convolutional layer, the middle layer is a max-over-time pooling, and the last layer is a fully connected output layer with a sigmoid activation function. With the CNN model, a category/label is given as output for any new unclassified text based on the experience derived from the training data. 

\subsection{Mapping result}
The mapping results derived from ES and CNN are represented as a dictionary of regulation controls with a corresponding confidence score. The confidence score measures the similarity between the techspec check input and the predicted regulation control(s).
In reality, techspec check mappings are often non-deterministic due to the inherent subjectivity in the mapping point of view. 
This subjectivity makes it imperative to find all regulation controls relevant to a given techspec check. To do this, the results of both CNN and ES are combined in a hybrid approach to capture more mapping possibilities while maintaining the relevance of the results. Example of mapping results are shown in Table \ref{tab:mapexamples}.


\subsection{Active Learning with SME Feedback}
It is difficult to achieve perfect mapping accuracy with any AI technique. {\name} addresses this by capturing continuous feedback from SMEs with a simple active learning mechanism where each additional sample is added to ES. CNN is also re-trained after every $y$ new sample where $y$ is a hyper-parameter, e.g., 50.

\section{Evaluation}
The performance of the proposed mapping algorithm is evaluated by running experiments with the dataset that is described next. 


    

\subsection{Dataset} \label{subsec:data}
We used a set of 429 STIGs documents containing $18757$ techspec checks for various technologies as training data for {\name}. Each security recommendation in these documents is mapped to one or more regulation controls in the NIST 800-53 v4 family. We combine the $title$, $description$, $rationale$, and $fix$ columns to form the specification text through the text pre-processing step and the corresponding NIST 800-53 v4 regulation control as the label. 
We divide the dataset into $k=3$ random folds with one fold used for testing and $k-1$ folds for training to initialize the algorithms. The testing fold accounts for 15\% of the dataset. Results presented are averaged over multiple iterations of the experiment.

\subsection{Experimental Result and Analysis}
\begin{figure}[h]
  \centering
  \includegraphics[width=1\linewidth]{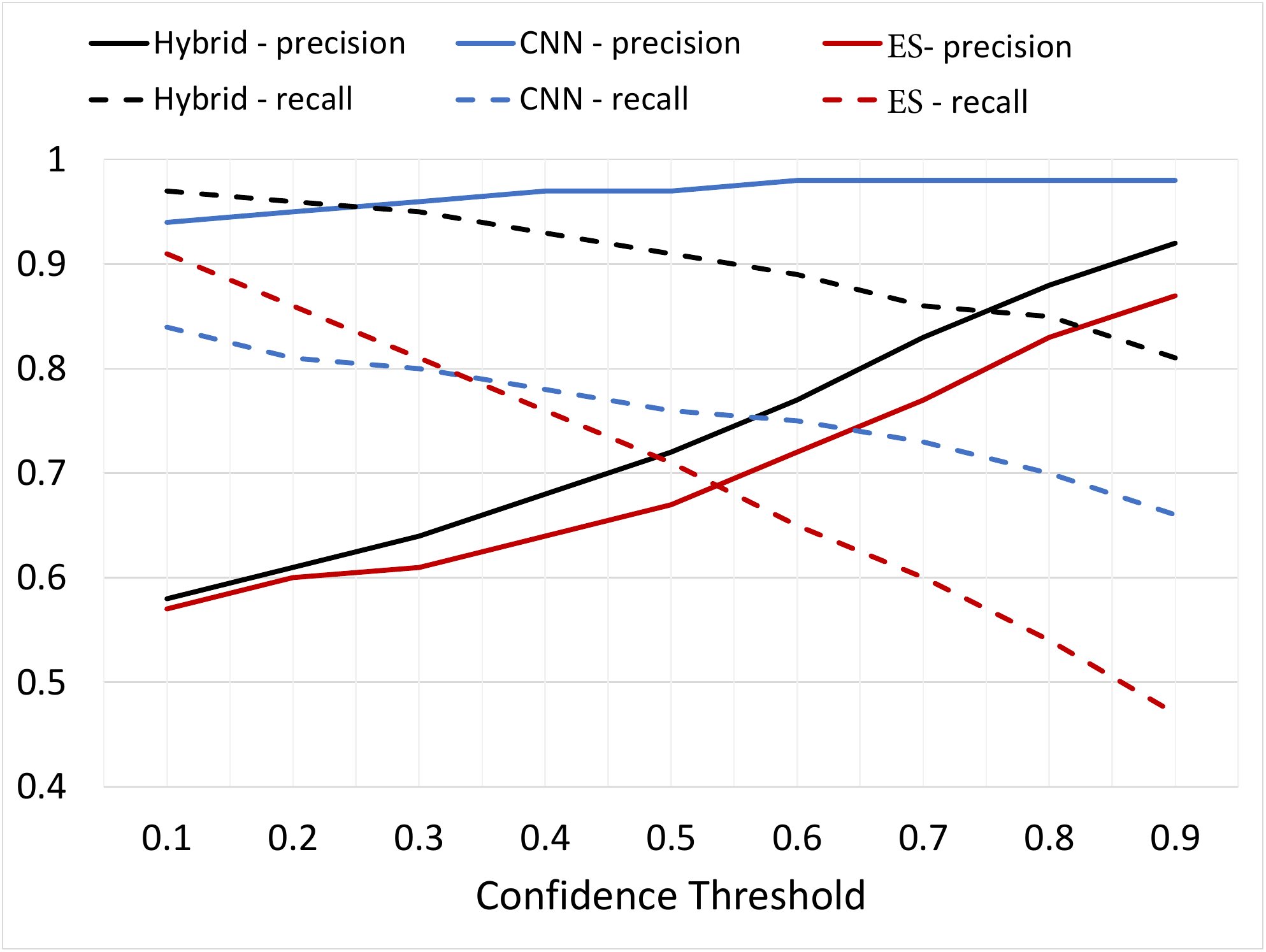}
  \caption{Performance analysis of different approaches on different confidence thresholds.}
  \label{fig:analysis}
\end{figure}

The first set of experiments were designed to prove the advantages of the hybrid proposed approach over the individual CNN and ES methodologies. We evaluated the results using the precision and recall metrics with varying confidence thresholds for the mapping result as shown in Figure \ref{fig:analysis}. Varying the confidence threshold provides an opportunity to evaluate the performance of the models based on the likelihood that the mapping result may be associated with the input techspec based on the training data. This provides an alternate approach to choosing the top $k$ results from the resultset. 

Across confidence thresholds, the Hybrid approach achieves a better recall than CNN and ES but is outperformed by CNN in terms of Precision. Based on the volume of training data and the structure, the CNN model reaches a tighter fit to the data hence the high precision. With the high recall, {\name} gives a better chance at retrieving all possible mappings for a given techspec check but also with a higher chance of having irrelevant mapping outcomes due to the lower precision relative to CNN. The hybrid approach however makes it easier for the SME to verify these mappings since it is often faster to identify incorrect mapping than it is to find correct ones. These verified mappings are subsequently used to improve the performance of {\name} in future iterations.
We also observe that at high confidence thresholds, the hybrid approach records precision closer to CNN, and achieves better recall than both CNN and ES. This allows the searcher to select higher confidence thresholds in {\name} without compromising precision and recall, unlike when using CNN or ES individually.

\begin{figure}[h]
  \centering
  \includegraphics[width=0.9\linewidth]{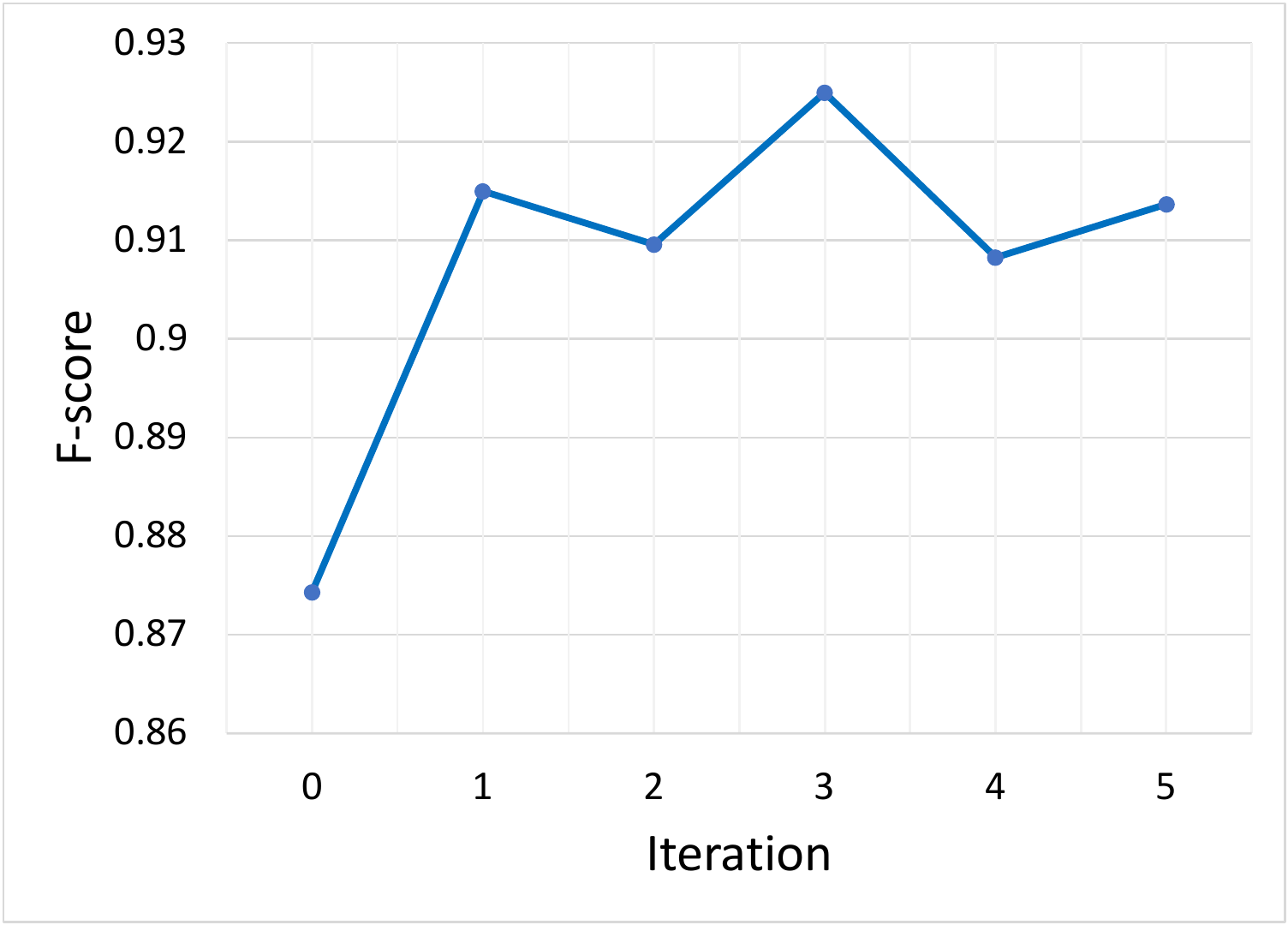}
  \caption{Effect of SME feedback on the accuracy of {\name}}
  \label{fig:feedback}
\end{figure}

To evaluate the impact of the active learning mechanism on {\name}, SME feedback is simulated by using a dataset of security rules from the RedHat Openshift Compliance Operator \cite{complianceoperator}. 360 security rules with existing mappings to NIST 800-53 v4 are selected to enrich the dataset in $5$ iterations of equal sizes such that $y=72$. $72$ security rules are added to X at each iteration, while CNN is also retrained with the additional rules. Iteration $0$ represents the performance of {\name} without SME feedback. 

As shown in Figure \ref{fig:feedback}, {\name} achieves better f-score with SME feedback ($\geq 0.908$). With no SME feedback captured, the f-score is ($0.874$). A significant improvement in the f-score is achieved after the first iteration, indicative of SME feedback's positive impact. We chose f-score as the evaluation metric as it provides us with an average view of the effect of the active learning process in reducing the number of incorrect mapping results, as well as in providing a good percentage of the actual correct mapping in the mapping result.  The cost of active learning by {\name} is also measured. Training the CNN model took an average of $0.08$ seconds/entry compared to the $2.2$ seconds/entry for data entry into ES.

\vspace{3mm}
\section{Challenges and Discussion}
\vspace{2mm}



\textbf{Level of details in text}: One of the major challenges of a mapping problem is the level of details in text. Regulations are usually written in \emph{legal} language with high level descriptions, whereas techspecs are written in e\emph{technical} language with low level details. This makes it challenging for AI techniques to match different granularity of information with each other. To alleviate the problem, as a future work, we plan to enrich the texts of regulations and techspecs with self-supervised transformer models to bring additional context in addition to the original content.\\
\\
\textbf{Weak labels for learning}: Keeping human in the loop is critical to the success of {\name}. Experts represent years of experience that compliments the AI approach of {\name}. However, the combination of skill sets (compliance and technical) required for experts in mapping is both unique and rare. Often this leads to inconsistencies in mappings, which in turn affects the accuracy of the AI model. {\name}'s methodology as an ensemble approach (Search + AI), to some extent, is a response to this problem. As a future work, we will extend this approach to rank experts and weight their contribution for better AI models. 



\section{Conclusion}

In this paper, we propose {\name} - an AI-assisted approach for mapping techspec checks to regulation controls with human in the loop. In future work, we will extend {\name} to achieve mapping to other regulatory standards, incorporate additional context by using transformer models and account for expertise difference of SMEs to achieve better accuracy. We will also expand the dataset to further validate the active learning accuracy.



\bibliographystyle{ACM-Reference-Format}
\bibliography{ref}


\begin{thebibliography}{28}


\ifx \showCODEN    \undefined \def \showCODEN     #1{\unskip}     \fi
\ifx \showDOI      \undefined \def \showDOI       #1{#1}\fi
\ifx \showISBNx    \undefined \def \showISBNx     #1{\unskip}     \fi
\ifx \showISBNxiii \undefined \def \showISBNxiii  #1{\unskip}     \fi
\ifx \showISSN     \undefined \def \showISSN      #1{\unskip}     \fi
\ifx \showLCCN     \undefined \def \showLCCN      #1{\unskip}     \fi
\ifx \shownote     \undefined \def \shownote      #1{#1}          \fi
\ifx \showarticletitle \undefined \def \showarticletitle #1{#1}   \fi
\ifx \showURL      \undefined \def \showURL       {\relax}        \fi
\providecommand\bibfield[2]{#2}
\providecommand\bibinfo[2]{#2}
\providecommand\natexlab[1]{#1}
\providecommand\showeprint[2][]{arXiv:#2}

\bibitem[\protect\citeauthoryear{Adam, Bulut, Hernandez, and Vukovic}{Adam
  et~al\mbox{.}}{2019}]%
        {adam2019cognitive}
\bibfield{author}{\bibinfo{person}{Constantin Adam},
  \bibinfo{person}{Muhammed~Fatih Bulut}, \bibinfo{person}{Milton Hernandez},
  {and} \bibinfo{person}{Maja Vukovic}.} \bibinfo{year}{2019}\natexlab{}.
\newblock \showarticletitle{Cognitive Compliance: Analyze, Monitor and Enforce
  Compliance in the Cloud}. In \bibinfo{booktitle}{\emph{2019 IEEE 12th
  International Conference on Cloud Computing (CLOUD)}}. IEEE,
  \bibinfo{pages}{234--242}.
\newblock


\bibitem[\protect\citeauthoryear{Agarwal, Bar-Haim, Eden, Gupta, Kantor, and
  Kumar}{Agarwal et~al\mbox{.}}{2021}]%
        {agarwal2021ai}
\bibfield{author}{\bibinfo{person}{Vikas Agarwal}, \bibinfo{person}{Roy
  Bar-Haim}, \bibinfo{person}{Lilach Eden}, \bibinfo{person}{Nisha Gupta},
  \bibinfo{person}{Yoav Kantor}, {and} \bibinfo{person}{Arun Kumar}.}
  \bibinfo{year}{2021}\natexlab{}.
\newblock \showarticletitle{AI-Assisted Security Controls Mapping for Clouds
  Built for Regulated Workloads}. In \bibinfo{booktitle}{\emph{2021 IEEE 14th
  International Conference on Cloud Computing (CLOUD)}}. IEEE,
  \bibinfo{pages}{136--146}.
\newblock


\bibitem[\protect\citeauthoryear{Berger}{Berger}{2015}]%
        {Berger2015LargeSM}
\bibfield{author}{\bibinfo{person}{Mark~J. Berger}.}
  \bibinfo{year}{2015}\natexlab{}.
\newblock \showarticletitle{Large Scale Multi-label Text Classification with
  Semantic Word Vectors}.
\newblock


\bibitem[\protect\citeauthoryear{CCPA}{CCPA}{2018}]%
        {ccpawebsite}
\bibfield{author}{\bibinfo{person}{CCPA}.} \bibinfo{year}{2018}\natexlab{}.
\newblock \bibinfo{booktitle}{\emph{CCPA Regulations}}.
\newblock
\urldef\tempurl%
\url{https://oag.ca.gov/privacy/ccpa}
\showURL{%
\tempurl}


\bibitem[\protect\citeauthoryear{Chalkidis, Fergadiotis, Malakasiotis, and
  Androutsopoulos}{Chalkidis et~al\mbox{.}}{2019}]%
        {chalkidis2019large}
\bibfield{author}{\bibinfo{person}{Ilias Chalkidis}, \bibinfo{person}{Manos
  Fergadiotis}, \bibinfo{person}{Prodromos Malakasiotis}, {and}
  \bibinfo{person}{Ion Androutsopoulos}.} \bibinfo{year}{2019}\natexlab{}.
\newblock \showarticletitle{Large-scale multi-label text classification on EU
  legislation}.
\newblock \bibinfo{journal}{\emph{arXiv preprint arXiv:1906.02192}}
  (\bibinfo{year}{2019}).
\newblock


\bibitem[\protect\citeauthoryear{Chen}{Chen}{2015}]%
        {chen2015convolutional}
\bibfield{author}{\bibinfo{person}{Yahui Chen}.}
  \bibinfo{year}{2015}\natexlab{}.
\newblock \emph{\bibinfo{title}{Convolutional neural network for sentence
  classification}}.
\newblock \bibinfo{thesistype}{Master's\ thesis}. \bibinfo{school}{University
  of Waterloo}.
\newblock


\bibitem[\protect\citeauthoryear{CIS}{CIS}{2021}]%
        {cis}
\bibfield{author}{\bibinfo{person}{CIS}.} \bibinfo{year}{2021}\natexlab{}.
\newblock \bibinfo{booktitle}{\emph{CIS Controls}}.
\newblock
\urldef\tempurl%
\url{https://www.cisecurity.org}
\showURL{%
\tempurl}


\bibitem[\protect\citeauthoryear{Devlin, Chang, Lee, and Toutanova}{Devlin
  et~al\mbox{.}}{2018}]%
        {devlin2018bert}
\bibfield{author}{\bibinfo{person}{Jacob Devlin}, \bibinfo{person}{Ming-Wei
  Chang}, \bibinfo{person}{Kenton Lee}, {and} \bibinfo{person}{Kristina
  Toutanova}.} \bibinfo{year}{2018}\natexlab{}.
\newblock \showarticletitle{Bert: Pre-training of deep bidirectional
  transformers for language understanding}.
\newblock \bibinfo{journal}{\emph{arXiv preprint arXiv:1810.04805}}
  (\bibinfo{year}{2018}).
\newblock


\bibitem[\protect\citeauthoryear{Forbes}{Forbes}{2018}]%
        {forbes}
\bibfield{author}{\bibinfo{person}{Forbes}.} \bibinfo{year}{2018}\natexlab{}.
\newblock \bibinfo{booktitle}{\emph{IBM's Big Bet On 'Hybrid' Cloud, Will It
  Work}}.
\newblock
\urldef\tempurl%
\url{https://www.forbes.com/sites/panosmourdoukoutas/2018/12/01/ibms-big-bet-on-hybrid-cloud-will-it-work/?sh=4853f62c734e}
\showURL{%
\tempurl}


\bibitem[\protect\citeauthoryear{GDPR}{GDPR}{2016}]%
        {gdprwebsite}
\bibfield{author}{\bibinfo{person}{GDPR}.} \bibinfo{year}{2016}\natexlab{}.
\newblock \bibinfo{booktitle}{\emph{General Data Protection Regulation (GDPR)
  Compliance Guidelines}}.
\newblock
\urldef\tempurl%
\url{https://gdpr.eu/}
\showURL{%
\tempurl}


\bibitem[\protect\citeauthoryear{GlobalNewsWire}{GlobalNewsWire}{2020}]%
        {newswire}
\bibfield{author}{\bibinfo{person}{GlobalNewsWire}.}
  \bibinfo{year}{2020}\natexlab{}.
\newblock \bibinfo{booktitle}{\emph{Cloud Computing Industry to Grow...}}
\newblock
\urldef\tempurl%
\url{https://www.globenewswire.com/news-release/2020/08/21/2081841/0/en/Cloud-Computing-Industry-to-Grow-from-371-4-Billion-in-2020-to-832-1-Billion-by-2025-at-a-CAGR-of-17-5.html}
\showURL{%
\tempurl}


\bibitem[\protect\citeauthoryear{HIPAA}{HIPAA}{2009}]%
        {hipaawebsite}
\bibfield{author}{\bibinfo{person}{HIPAA}.} \bibinfo{year}{2009}\natexlab{}.
\newblock \bibinfo{booktitle}{\emph{Health insurance portability and
  accountability act - the security rule}}.
\newblock
\urldef\tempurl%
\url{https://www.hhs.gov/hipaa/forprofessionals/security/index.html}
\showURL{%
\tempurl}


\bibitem[\protect\citeauthoryear{Hughes, Li, Kotoulas, and Suzumura}{Hughes
  et~al\mbox{.}}{2017}]%
        {hughes2017medical}
\bibfield{author}{\bibinfo{person}{Mark Hughes}, \bibinfo{person}{Irene Li},
  \bibinfo{person}{Spyros Kotoulas}, {and} \bibinfo{person}{Toyotaro
  Suzumura}.} \bibinfo{year}{2017}\natexlab{}.
\newblock \showarticletitle{Medical text classification using convolutional
  neural networks}.
\newblock In \bibinfo{booktitle}{\emph{Informatics for Health: Connected
  Citizen-Led Wellness and Population Health}}. \bibinfo{publisher}{IOS Press},
  \bibinfo{pages}{246--250}.
\newblock


\bibitem[\protect\citeauthoryear{Jabreel and Moreno}{Jabreel and
  Moreno}{2019}]%
        {jabreel2019deep}
\bibfield{author}{\bibinfo{person}{Mohammed Jabreel} {and}
  \bibinfo{person}{Antonio Moreno}.} \bibinfo{year}{2019}\natexlab{}.
\newblock \showarticletitle{A deep learning-based approach for multi-label
  emotion classification in tweets}.
\newblock \bibinfo{journal}{\emph{Applied Sciences}} \bibinfo{volume}{9},
  \bibinfo{number}{6} (\bibinfo{year}{2019}), \bibinfo{pages}{1123}.
\newblock


\bibitem[\protect\citeauthoryear{Kim}{Kim}{2014}]%
        {Kim14f}
\bibfield{author}{\bibinfo{person}{Yoon Kim}.} \bibinfo{year}{2014}\natexlab{}.
\newblock \showarticletitle{Convolutional Neural Networks for Sentence
  Classification}.
\newblock \bibinfo{journal}{\emph{CoRR}}  \bibinfo{volume}{abs/1408.5882}
  (\bibinfo{year}{2014}).
\newblock
\showeprint[arxiv]{1408.5882}


\bibitem[\protect\citeauthoryear{Lan, Chen, Goodman, Gimpel, Sharma, and
  Soricut}{Lan et~al\mbox{.}}{2019}]%
        {lan2019albert}
\bibfield{author}{\bibinfo{person}{Zhenzhong Lan}, \bibinfo{person}{Mingda
  Chen}, \bibinfo{person}{Sebastian Goodman}, \bibinfo{person}{Kevin Gimpel},
  \bibinfo{person}{Piyush Sharma}, {and} \bibinfo{person}{Radu Soricut}.}
  \bibinfo{year}{2019}\natexlab{}.
\newblock \showarticletitle{Albert: A lite bert for self-supervised learning of
  language representations}.
\newblock \bibinfo{journal}{\emph{arXiv preprint arXiv:1909.11942}}
  (\bibinfo{year}{2019}).
\newblock


\bibitem[\protect\citeauthoryear{Liu, Chang, Wu, and Yang}{Liu
  et~al\mbox{.}}{2017}]%
        {liu2017deep}
\bibfield{author}{\bibinfo{person}{Jingzhou Liu}, \bibinfo{person}{Wei-Cheng
  Chang}, \bibinfo{person}{Yuexin Wu}, {and} \bibinfo{person}{Yiming Yang}.}
  \bibinfo{year}{2017}\natexlab{}.
\newblock \showarticletitle{Deep learning for extreme multi-label text
  classification}. In \bibinfo{booktitle}{\emph{Proceedings of the 40th
  International ACM SIGIR Conference on Research and Development in Information
  Retrieval}}. \bibinfo{pages}{115--124}.
\newblock


\bibitem[\protect\citeauthoryear{Liu, Ott, Goyal, Du, Joshi, Chen, Levy, Lewis,
  Zettlemoyer, and Stoyanov}{Liu et~al\mbox{.}}{2019}]%
        {liu2019roberta}
\bibfield{author}{\bibinfo{person}{Yinhan Liu}, \bibinfo{person}{Myle Ott},
  \bibinfo{person}{Naman Goyal}, \bibinfo{person}{Jingfei Du},
  \bibinfo{person}{Mandar Joshi}, \bibinfo{person}{Danqi Chen},
  \bibinfo{person}{Omer Levy}, \bibinfo{person}{Mike Lewis},
  \bibinfo{person}{Luke Zettlemoyer}, {and} \bibinfo{person}{Veselin
  Stoyanov}.} \bibinfo{year}{2019}\natexlab{}.
\newblock \showarticletitle{Roberta: A robustly optimized bert pretraining
  approach}.
\newblock \bibinfo{journal}{\emph{arXiv preprint arXiv:1907.11692}}
  (\bibinfo{year}{2019}).
\newblock


\bibitem[\protect\citeauthoryear{magpie}{magpie}{2018}]%
        {magpie}
\bibfield{author}{\bibinfo{person}{magpie}.} \bibinfo{year}{2018}\natexlab{}.
\newblock \bibinfo{booktitle}{\emph{Deep neural network framework for
  multilabel text classification}}.
\newblock
\urldef\tempurl%
\url{https://github.com/inspirehep/magpie}
\showURL{%
\tempurl}


\bibitem[\protect\citeauthoryear{NV}{NV}{2022}]%
        {elastic}
\bibfield{author}{\bibinfo{person}{Elastic NV}.}
  \bibinfo{year}{2022}\natexlab{}.
\newblock \bibinfo{booktitle}{\emph{Elasticsearch}}.
\newblock
\urldef\tempurl%
\url{https://github.com/elastic/elasticsearch}
\showURL{%
\tempurl}


\bibitem[\protect\citeauthoryear{Openshift}{Openshift}{2020}]%
        {complianceoperator}
\bibfield{author}{\bibinfo{person}{Openshift}.}
  \bibinfo{year}{2020}\natexlab{}.
\newblock \bibinfo{booktitle}{\emph{Compliance Operator}}.
\newblock
\urldef\tempurl%
\url{https://github.com/openshift/compliance-operator}
\showURL{%
\tempurl}


\bibitem[\protect\citeauthoryear{PCI-DSS}{PCI-DSS}{2016}]%
        {pcidsswebsite}
\bibfield{author}{\bibinfo{person}{PCI-DSS}.} \bibinfo{year}{2016}\natexlab{}.
\newblock \bibinfo{booktitle}{\emph{Payment card industry (pci) data security
  standard}}.
\newblock
\urldef\tempurl%
\url{https://www.pcisecuritystandards.org/}
\showURL{%
\tempurl}


\bibitem[\protect\citeauthoryear{STIGS}{STIGS}{2018}]%
        {stigs}
\bibfield{author}{\bibinfo{person}{STIGS}.} \bibinfo{year}{2018}\natexlab{}.
\newblock \bibinfo{booktitle}{\emph{Security technical implementation guides}}.
\newblock
\urldef\tempurl%
\url{https://https://public.cyber.mil/stigs}
\showURL{%
\tempurl}


\bibitem[\protect\citeauthoryear{Tay, Dehghani, Gupta, Bahri, Aribandi, Qin,
  and Metzler}{Tay et~al\mbox{.}}{2021}]%
        {tay2021pre}
\bibfield{author}{\bibinfo{person}{Yi Tay}, \bibinfo{person}{Mostafa Dehghani},
  \bibinfo{person}{Jai Gupta}, \bibinfo{person}{Dara Bahri},
  \bibinfo{person}{Vamsi Aribandi}, \bibinfo{person}{Zhen Qin}, {and}
  \bibinfo{person}{Donald Metzler}.} \bibinfo{year}{2021}\natexlab{}.
\newblock \showarticletitle{Are pre-trained convolutions better than
  pre-trained transformers?}
\newblock \bibinfo{journal}{\emph{arXiv preprint arXiv:2105.03322}}
  (\bibinfo{year}{2021}).
\newblock


\bibitem[\protect\citeauthoryear{You, Zhang, Wang, Dai, Mamitsuka, and Zhu}{You
  et~al\mbox{.}}{2018}]%
        {you2018attentionxml}
\bibfield{author}{\bibinfo{person}{Ronghui You}, \bibinfo{person}{Zihan Zhang},
  \bibinfo{person}{Ziye Wang}, \bibinfo{person}{Suyang Dai},
  \bibinfo{person}{Hiroshi Mamitsuka}, {and} \bibinfo{person}{Shanfeng Zhu}.}
  \bibinfo{year}{2018}\natexlab{}.
\newblock \showarticletitle{Attentionxml: Label tree-based attention-aware deep
  model for high-performance extreme multi-label text classification}.
\newblock \bibinfo{journal}{\emph{arXiv preprint arXiv:1811.01727}}
  (\bibinfo{year}{2018}).
\newblock


\bibitem[\protect\citeauthoryear{Zhang, Zhao, and LeCun}{Zhang
  et~al\mbox{.}}{2015}]%
        {zhang2015character}
\bibfield{author}{\bibinfo{person}{Xiang Zhang}, \bibinfo{person}{Junbo Zhao},
  {and} \bibinfo{person}{Yann LeCun}.} \bibinfo{year}{2015}\natexlab{}.
\newblock \showarticletitle{Character-level convolutional networks for text
  classification}.
\newblock \bibinfo{journal}{\emph{Advances in neural information processing
  systems}}  \bibinfo{volume}{28} (\bibinfo{year}{2015}).
\newblock


\bibitem[\protect\citeauthoryear{Zhou, Sun, Liu, and Lau}{Zhou
  et~al\mbox{.}}{2015}]%
        {zhou2015c}
\bibfield{author}{\bibinfo{person}{Chunting Zhou}, \bibinfo{person}{Chonglin
  Sun}, \bibinfo{person}{Zhiyuan Liu}, {and} \bibinfo{person}{Francis Lau}.}
  \bibinfo{year}{2015}\natexlab{}.
\newblock \showarticletitle{A C-LSTM neural network for text classification}.
\newblock \bibinfo{journal}{\emph{arXiv preprint arXiv:1511.08630}}
  (\bibinfo{year}{2015}).
\newblock


\bibitem[\protect\citeauthoryear{Zhou, Qi, Zheng, Xu, Bao, and Xu}{Zhou
  et~al\mbox{.}}{2016}]%
        {zhou2016text}
\bibfield{author}{\bibinfo{person}{Peng Zhou}, \bibinfo{person}{Zhenyu Qi},
  \bibinfo{person}{Suncong Zheng}, \bibinfo{person}{Jiaming Xu},
  \bibinfo{person}{Hongyun Bao}, {and} \bibinfo{person}{Bo Xu}.}
  \bibinfo{year}{2016}\natexlab{}.
\newblock \showarticletitle{Text classification improved by integrating
  bidirectional LSTM with two-dimensional max pooling}.
\newblock \bibinfo{journal}{\emph{arXiv preprint arXiv:1611.06639}}
  (\bibinfo{year}{2016}).
\newblock


\end{thebibliography}

\end{document}